\documentclass[conference]{IEEEtran}
\usepackage[cmex10]{amsmath}
\usepackage{times}
\usepackage{epsfig}
\usepackage{graphicx}
\usepackage{amsmath}
\usepackage{amssymb}
\DeclareMathOperator{\E}{\mathbb{E}}
\usepackage[utf8]{inputenc}
\usepackage{cite}
\usepackage{wrapfig}
\usepackage{float}
\usepackage{tabularx}
\usepackage{booktabs}
\usepackage[colorlinks,allcolors=red]{hyperref}
\usepackage{url}
\usepackage{fixltx2e}
\usepackage{stfloats}
\usepackage{fancyhdr}
\begin{document}
\pagestyle{fancy}
\title{PerceptionGAN: Real-world Image Construction from Provided Text through Perceptual Understanding}

\author{\IEEEauthorblockN{{Kanish Garg}\IEEEauthorrefmark{1},
{Ajeet kumar Singh}\IEEEauthorrefmark{1},
{Dorien Herremans}\IEEEauthorrefmark{2} and
{Brejesh Lall}\IEEEauthorrefmark{1}}

\IEEEauthorblockA{\IEEEauthorrefmark{1} Indian Institute of Technology Delhi, India, \emph{\{kanishgarg428, ajeetsngh24\}@gmail.com , brejesh@ee.iitd.ac.in}
}

\IEEEauthorblockA{\IEEEauthorrefmark{2} Singapore University of Technology and Design, Singapore, \emph{dorien\_herremans@sutd.edu.sg}
}
% \IEEEauthorblockA{
% \{kanishgarg428,
% ajeetsngh24\}@gmail.com,
% dorien\_herremans@sutd.edu.sg,
% brejesh@ee.iitd.ac.in
% }
}

\maketitle
\thispagestyle{fancy}

\begin{abstract}
% \boldmath
Generating an image from a provided descriptive text is quite a challenging task because of the difficulty in incorporating perceptual information (object shapes, colors, and their interactions) along with providing high relevancy related to the provided text. Current methods first generate an initial low-resolution image, which typically has irregular object shapes, colors, and interaction between objects. This initial image is then improved by conditioning on the text. However, these methods mainly address the problem of using text representation efficiently in the refinement of the initially generated image, while the success of this refinement process depends heavily on the quality of the initially generated image, as pointed out in the Dynamic Memory Generative Adversarial Network (DM-GAN) paper. Hence, we propose a method to provide good initialized images by incorporating perceptual understanding in the discriminator module. We improve the perceptual information at the first stage itself, which results in significant improvement in the final generated image. In this paper, we have applied our approach to the novel StackGAN architecture. We then show that the perceptual information included in the initial image is improved while modeling image distribution at multiple stages. Finally, we generated realistic multi-colored images conditioned by text. These images have good quality along with containing improved basic perceptual information. More importantly, the proposed method can be integrated into the pipeline of other state-of-the-art text-based-image-generation models such as DM-GAN and AttnGAN to generate initial low-resolution images. We also worked on improving the refinement process in StackGAN by augmenting the third stage of the generator-discriminator pair in the StackGAN architecture. Our experimental analysis and comparison with the state-of-the-art on a large but sparse dataset MS COCO further validate the usefulness of our proposed approach.
\end{abstract}
\vspace{8pt}
\textbf{Contribution--}This paper improves the pipeline for Text to Image Generation by incorporating Perceptual Understanding in the Initial Stage of Image Generation.


\pagestyle{fancy}
\section{\textbf{Introduction}}
Generating photo-realistic images from unstructured text descriptions is a very challenging problem in computer vision but has many potential applications in the real world. Over the past years, with the advances in the meaningful representation of the text (text-embeddings) through the use of extremely powerful models such as word2vec, GloVe~\cite{28,29,30} combined with RNNs, multiple architectures have been proposed for image retrieval and generation. Generative Adversarial Networks (GANs)~\cite{3} achieved significant results and have gained a lot of attention in regards to text to image synthesis. As discussed in StackGAN paper~\cite{4}, natural images can be modeled at different scales; hence GANs can be stably trained for multiple sub-generative tasks with progressive goals. Thus current state-of-the-art methods generate images by modeling a series of low-to-high-dimensional data distributions, which can be viewed as first generating low-resolution images with basic object shapes, colors, and then converting these images to high-resolution ones. However, there are two major problems that need to be addressed~\cite{1}. 1) Existing methods depend heavily on the quality of the initially generated image, and if this is not well initialized (i.e., not able to capture basic object shapes, colors, and interactions between objects), then further refinement will not be able to improve the quality much. 2) Each word provides a contribution of different levels of importance when depicting different image content; hence unchanged word embeddings in the refinement process make the final model less effective. Most of the current state-of-the-art methods have only addressed the second problem~\cite{1,2,4}. In contrast, in this paper, we propose a novel method to address the first problem, namely, generating a good, perceptually relevant, low-resolution image to be used as an initialization for the refinement stage.
\\
In the first step of our research, we tried to improve the refinement process by augmenting the third stage of the generator-discriminator pair in the StackGAN architecture. Based on the analysis of the results, unfortunately, adding more stages in the refinement process did not significantly improve perceptual aspects like object shapes, object interactions, etc. in the final generated image compared with the cost of the increase in parameters. Based on this preliminary research, we project that the presence of perceptual aspects should be addressed in the first stage of the generation itself. Hence, we propose a method to provide good initialized images by incorporating perceptual understanding in the discriminator module. We introduced an encoder in the discriminator module of the first stage. The encoder maps the image distribution to a lower-dimensional distribution, which still contains all the relevant perceptual information. A schematic overview of the mappings in our proposed PerceptionGAN architecture is shown in \textbf{Fig. 1.}. We introduce a captioner loss on this lower-dimensional vector of the real and generated image. This ensures that the generated image contains most of the relevant perceptual aspects present in the input training image in the first stage of the generation itself, which will enhance any further refinement process that aims to improve the quality of the text conditioned image.
\\
\begin{figure}[h]
\centering
\includegraphics[width=7 cm]{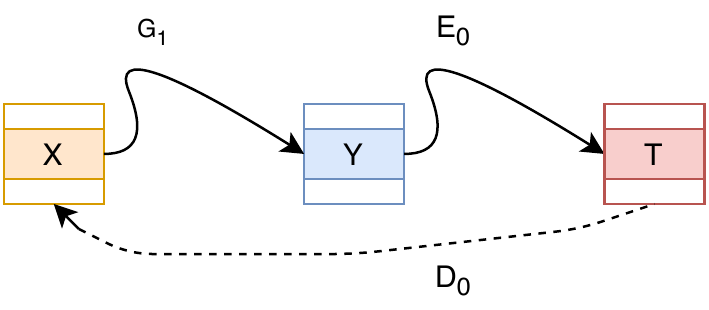}
\label{mappings}
\caption{A schematic overview of the mappings in our proposed PerceptionGAN: $ G_1: X \to Y$ and $ E_0: Y \to T$, where X represents the text representation space, and Y represents the image space. T is a 256-dimensional latent space-constrained such that there exists a reverse mapping, i.e., a decoder $D_0: T \to X$   }
\end{figure}

\section{\textbf{Related Work}}
Generating high-resolution images from unstructured text is a difficult task and is very useful in the field of computer-aided design, text conditioned generation of images of objects, etc. Several deep generative models have been designed for the synthesis of images from unstructured text representations. The AlignDRAW model by Mansimov et al.~\cite{5} iteratively draws patches on a canvas while attending to the relevant words in the description. The Conditional PixelCNN used by Reed et al.~\cite{31} uses text descriptors as conditional variables. The Approximate Langevin sampling approach was used by Nguyen et al.~\cite{7} to generate images conditioned on the text. Compared to other generative models, however, GANs have shown a much better performance when it comes to image synthesis from text. In particular, Conditional Generative Networks have achieved significant results in this domain.
Reed et al.~\cite{13} successfully generated $64\times$64 resolution images for birds and flowers using text descriptions with GANs. These generated samples were further improved by considering additional information regarding object location in their following research~\cite{6}. To capture the plenitude of information present in natural images, several multiple-GAN architectures were also proposed. Wang et al.~\cite{8} utilized a structure GAN and a style GAN to synthesize images of indoor scenes. Yang et al.~\cite{9} used layered recursive GANs to factorize image generation into foreground and background generation. Several GANs were added by Huang et al.~\cite{10} to reconstruct the multilevel representations of a pre-trained discriminative model. Durugkar et al.~\cite{11} used multiple discriminators along with one generator to increase the probability of the generator acquiring effective feedback. This approach, however, lacks in modeling image distribution at multiple discrete scales. Denton et al.~\cite{12} built a series of GANs within a Laplacian pyramid framework (LAPGANs) where a residual image conditioned on the image of the previous stage is generated and then added back to the input image to produce the input for the next stage. Han Zhang and Tao Xu's work on StackGAN~\cite{4} and StackGAN++~\cite{14} further improved the final image quality and relevancy. The latter work included various features (Conditioning Augmentation, Color Consistency regulation), which led to a further improvement in image generation. AttnGAN~\cite{2} also achieved good results in this field. The idea behind AttnGAN is to refine the images to high-resolution ones by leveraging the attention mechanism. Each word in an input sentence has a different level of information depicting the image content. So instead of conditioning images on global sentence vectors, they conditioned images on fine-grained word-level information, during which they consider all of the words equally. Dynamic Memory GAN (DM-GAN)~\cite{1} improved the word selection mechanism for image generation, by dynamically selecting the important word information based on the initially generated image and then refining the image conditioned on that information part by part. These models, unfortunately, are mostly targeting the problem of incorporating the provided text description more efficiently into the refinement stage of the low-resolution image and do not address the problem of lacking perceptual information in the initially generated low-resolution image. As observed in all of the models above, they are only able to capture a limited amount of perceptual information, i.e., the final generated image is not photo-realistic. Hence we project that it is important to capture perceptual information in the first stage itself. Our PerceptionGAN architecture targets this important problem and improves the perceptual information in the initially generated image significantly, which is then even further improved in the refinement process. Our proposed PerceptionGAN approach, not only approximates image distribution at multiple discrete scales by generating high-resolution images conditioned on their low-resolution counterparts (generated in the initial stage); as done similarly in StackGAN, StackGAN++, LAPGANs, Progressive GANs; they also offer a major improvement of final image quality by creating the initial low-resolution input images through the incorporation of perceptual content. Furthermore, our PerceptionGAN increases the text relevancy of the images through a loss analysis between the perceptual features' distribution of generated and real images.

\section{\textbf{Proposed Architecture}}
As stated by Zhu, Minfeng, et al.~\cite{1}, who proposed DM-GAN, and as verified in our preliminary experiment below, the first stage of image generation needs to capture more perceptual information to improve the final output. Hence, we propose a new architecture, shown in \textbf{Fig. 2.}, in which we introduce an encoder in the discriminator module of the first stage. This encoder maps the image distribution to a lower-dimensional distribution, which contains all of the relevant perceptual information. A captioner loss on this lower-dimensional vector of both the real and generated image is introduced along with the adversarial loss. This ensures the presence of most of the relevant perceptual aspects in the first stage of the generation itself, which will enhance any further refinement process that aims to improve the quality of the text conditioned image.

\begin{figure*}
\centering
\includegraphics[width=16 cm]{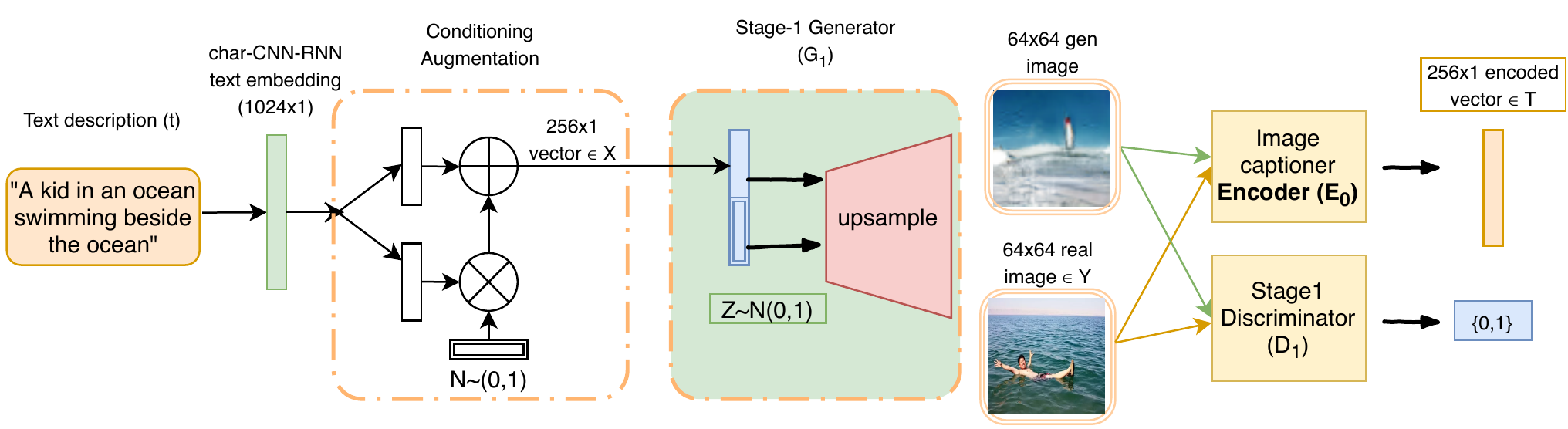}
\label{pgan}
\caption{Our proposed PerceptionGAN Architecture: An image captioner encoder ($ E_0 : Y \to T$) is introduced along with the stage-I discriminator ($D_1$). t is an input text description which is encoded to embeddings using a pre-trained char-CNN-RNN encoder~\cite{39}. Conditional Augmentation~\cite{4} is applied on the embeddings and the resulting text representation vector is passed through a generator $G_1 : X \to Y$ to generate images. $G_1$ consists of a series of upsampling and residual blocks. The layers of $G_1$ and $D_1$ are same as used in StackGAN paper. The architecture of image captioner encoder is shown in \textbf{Fig. 3.}}
\end{figure*}

\subsection{\textbf{Encoder}}
The encoder shown in \textbf{Fig. 3.} is the most important part of our proposed architecture. Our goal is to learn a mapping, as shown in \textbf{Fig. 1.}, between two domains X (text representation space) and Y (image space) given $N$ training samples $\{x_{i}\} _{i=1} ^{N} \in X$ with labels $\{y_{j}\} _{j=1} ^{N} \in Y $. The role of the encoder ($E_0$) is to map the high dimensional (64x64) image distribution to a low dimensional (256x1) distribution, i.e. $E_0 : Y \to T $  where T (with distribution $p_t$ ) is such that all of the relevant perceptual information is preserved. To ensure this, $p_t$  has to be such that the reverse mapping $D_0 : T \to X$ is attained with a decoder $D_0$. We trained an image captioner~\cite{25} and used its encoder ($E_0$) (with latent space of 256x1) for this purpose.

\begin{figure*}
\centering
\includegraphics[width=17 cm]{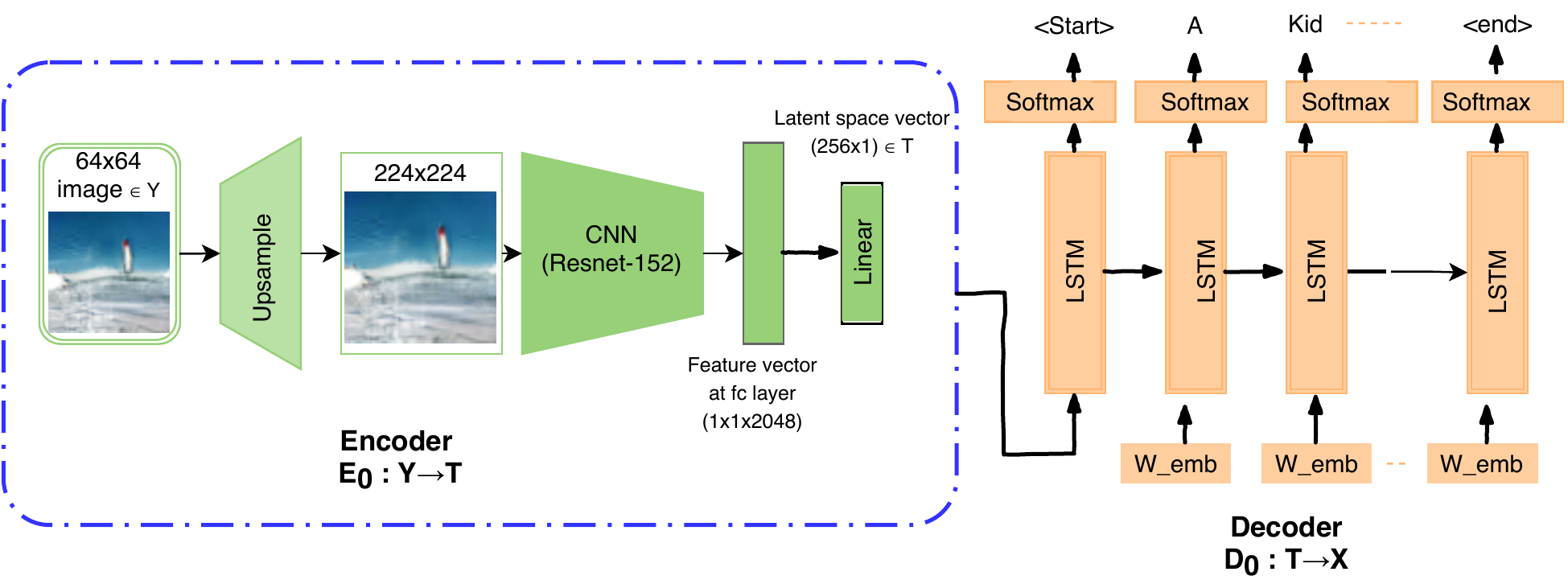}
\label{encoder}
\caption{Image captioner architecture, consisting of an encoder ($ E_0 : Y \to T$) and decoder ($ D_0 : T \to X$). We trained this Image captioner architecture end-to-end and used its encoder ($E_0$) as shown in \textbf{Fig. 2.}}
\end{figure*}

Now, given X and Y, we train a $G_1$ such that the total loss is minimized. Here, the total loss is defined as the Captioner Loss plus the Adversarial loss.

\subsection{\textbf{Adversarial Loss}}
For the mapping function $G_1 : X \to Y$ and its discriminator $D_1$, the adversarial objective $L_{GAN}$~\cite{3} can be expressed as:\\
\begin{multline}
L_{GAN}(G_1, D_1, X, Y) = \E_{y \sim p_{data}}[log D_1(y)] +
\\ \hfill \E_{x \sim p_{data}}[log(1 - D_1(G_1(x)))]]
\end{multline}

Where $G_1$ tries to generate images $G_1(x)$ that look similar to images from the domain Y, while $D_1$ aims to distinguish between the translated samples $G_1(x)$ and real samples. Generator $G_1$ tries to minimize the objective against the Discriminator $D_1$ that tries to maximize it. 

\subsection{\textbf{Captioner Loss}}
Adversarial training can, in theory, learn a mapping $G_1$, which produces outputs that are distributed identically as in the target domain Y~\cite{3}. The training images, however, also contain a lot of irrelevant and variable information other than the objects, their shapes, colors, interactions, etc., which makes training challenging.

To overcome this problem, we introduce a Captioner loss, which can account for loss related to perceptual features like shapes, object interactions, etc. This ensures that most of the perceptual information is extracted in the first stage before any further refinement is applied to improve the final image quality. This Captioner loss is the mean squared error (MSE) loss between the encoded vectors of real and generated images. The encoding is a one to one mapping between the high dimensional image space and the low-dimensional latent space containing all of the relevant perceptual information. The objective function can be written as: 
\begin{multline}
L_{Captioner}(G_1, X, Y)=
\\ \hfill \E_{x,y \sim p_{data}}[ MSE(E_0(y)-E_0(G_1(x)))] 
\end{multline}
\subsection{\textbf{Training details}}
The training process of GANs can be unstable due to multiple reasons~\cite{33,34,35} and is extremely dependent on the hyper-parameters tuning. We could argue that the disjoint nature in the data distribution and the corresponding model distribution may be one of the reasons for this instability. This problem becomes more apparent when GANs are used to generate high-resolution images because it further reduces the chance of a share of support between model and image distributions~\cite{4}. GANs can be trained in a stable manner~\cite{18} to generate high-resolution images that are conditioned on provided text description because the chance of sharing support in high dimensional space can be expected in the conditional distribution.

Furthermore, to make the training process more stable, we introduced the encoder after a certain number of epochs (5 epochs, chosen heuristically) of training (If the encoder is added right after the start of the training of Generator-Discriminator, it could contribute to a huge loss resulting in instability of the training process). This ensures that at the time encoder is introduced, the generated image will be having some rough object shapes, colors, their interactions, and massive loss will not be expected to occur, and the vanishing gradient problem will be checked.

\begin{figure}[ht]
\centering
\includegraphics[width=9 cm]{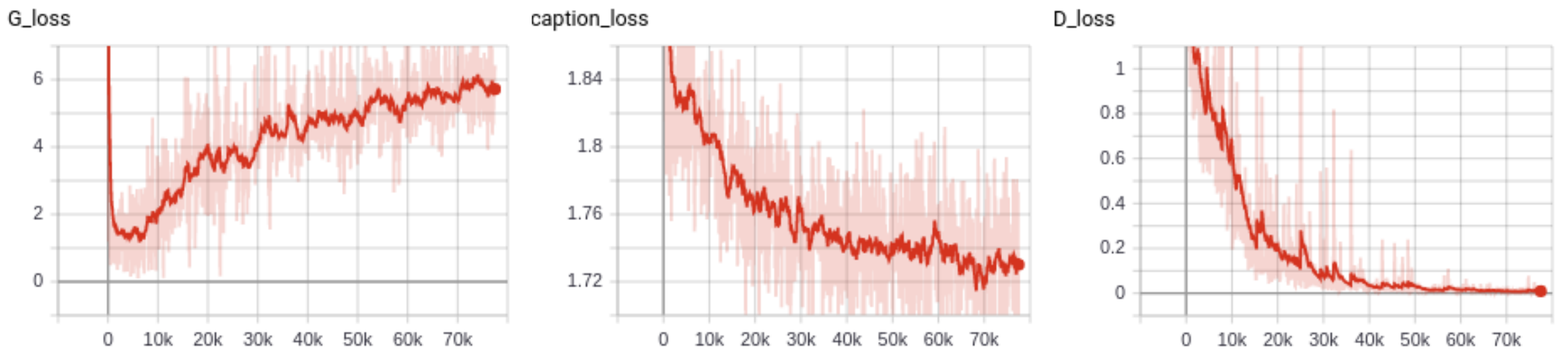}
\label{fig:learning}
\caption{ The evolution of the - Column 1: Generator loss; Column 2: Captioner loss; Column 3: Discriminator loss; during training. }
\end{figure}

We first trained the image captioner model, i.e., $E_0$ and $D_0$ 
and then use $E_0$ as a pretrained mapping in our PerceptionGAN architecture. 
We trained our PerceptionGAN on the MS-COCO dataset~\cite{26} (80,000 images and five descriptions per image) for 90 epochs with a batch size of 128. An NVIDIA Quadro P5000 with 16 GB GDDR5 memory and 2560 Cuda cores was used for training.

%---------------------------------------------------------------
\section{\textbf{Preliminary experiment to improve the refinement process of StackGAN}}
To enhance the refinement process of StackGAN and to observe how much perceptual information is included in the generated image by adding more stages in the refinement process, we added a third stage in the architecture of StackGAN. 

\subsection{\textbf{Added Stage-III GAN}}

The proposed third stage is similar to the second stage in the architecture of StackGAN as it repeats the conditioning process which helps the third stage acquire features omitted by the previous stages. The Stage-III GAN generates a high resolution image by conditioning on both of the previously generated image and text embedding vector \^{c}\begin{math} _{3}\end{math} and hence takes into account the text features to correct the defects in the image. In this stage, the discriminator $D_{3}$ and generator $G_{3}$ are trained by minimizing $-\alpha D_{3}$ (Discriminator Loss) and $\alpha G_{3}$ (Generator Loss) alternatively by conditioning on the previously generated image $s_{2} = G_{2}(s_{1}$, \^{c}$ _{2}$) and Gaussian latent variables \^{c}$ _{3}$. 
\begin{multline}
\alpha D_{3} = \E_{(y,t)\sim p_{data}} [log D_{3}(y,\varphi_{t})] +\\ \E _{s_{2}\sim p_{g_{2}},t\sim p_{data}} [log(1 - D_{3}(G_{3}(s_{2}, \text{\^{c}}_{3}),\varphi_{t}))]
\end{multline}
\begin{multline}
\alpha G_{3}= \E _{s_{2} \sim p_{g_{2}},t\sim p_{data}} [log(1-D_{3}(G_{3}(s_{2}, \text{\^{c}}_{3}), \varphi_{t}))] +\\\lambda D_{KL}( \mathcal{N}(\mu_{3}(\varphi_{t}), \Sigma_{3}(\varphi_{t}))|| \mathcal{N} (0, I)) 
\end{multline}
\begin{figure}[h!]
\includegraphics[width=3.4in,height=2.5in,keepaspectratio]{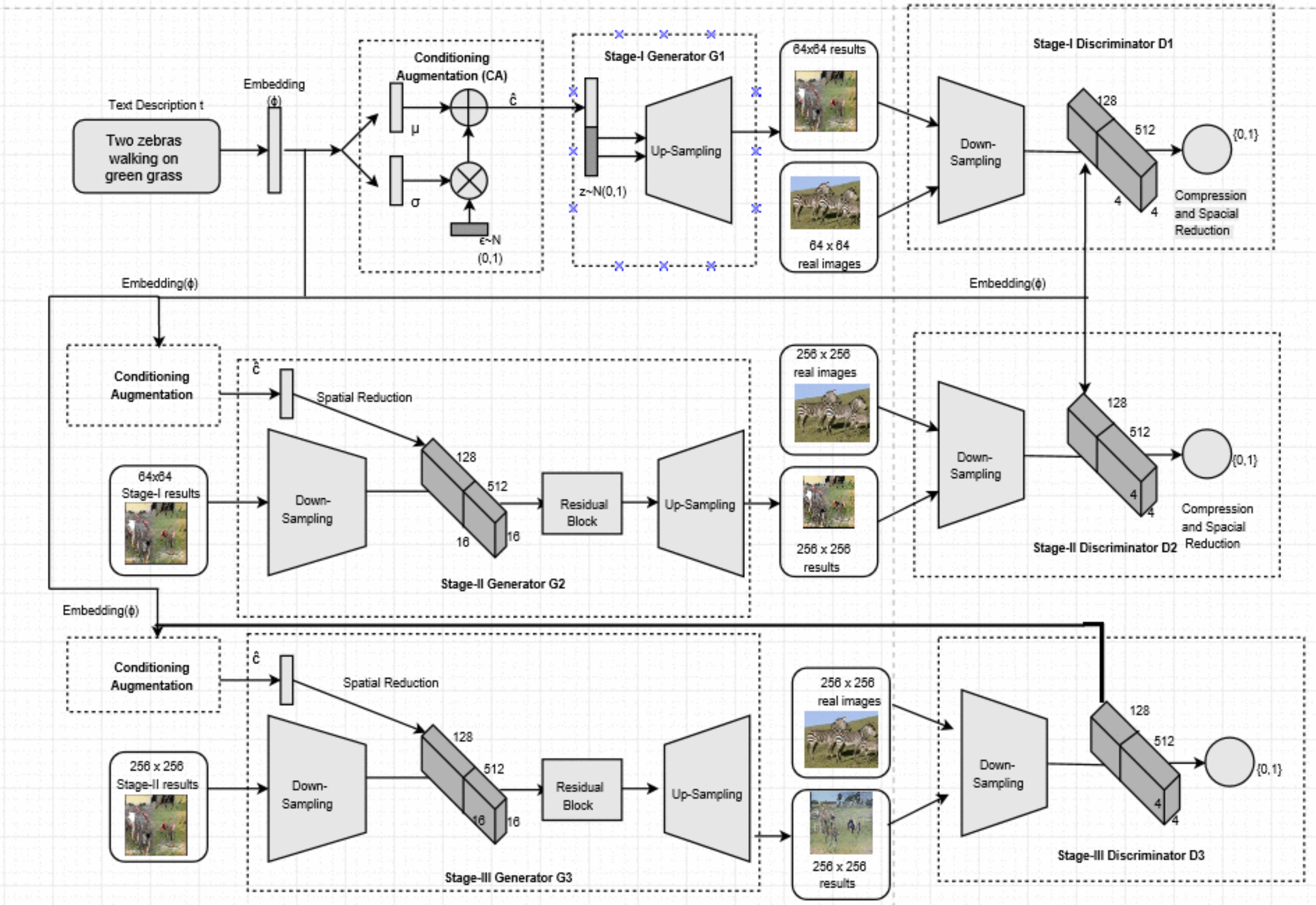}
\label{fig:stage3}
\caption{A third Stage is added in our novel architecture of StackGAN to observe how much perceptual information is improving in the finally generated image. The architecture of the Stage-III is kept similar to that of Stage-II. (Full-resolution picture is available online: \href{https://iitd.info/stage3}{https://iitd.info/stage3})}
\end{figure}
\textbf{Stage-III Architecture details :}
The pre-trained char-RNN-CNN text encoder~\cite{39} generates the text embeddings \begin{math}\varphi_{t}\end{math} as in Stage-I and Stage-II. However, since for both stages, different means and standard deviations are generated by the fully connected layers in the Conditioning Augmentation process~\cite{4}, the Stage-III GAN learns information which is omitted by the previous stage. The Stage-III generator is created as an encoder-decoder network that contains residual blocks~\cite{37}. The \begin{math}N_{g}\end{math} dimensional text conditioning vector \^{c}\begin{math}_{3}\end{math} is created using the char-CNN-RNN text embedding \begin{math}\varphi_{t}\end{math}, which is spatially replicated to form a M\begin{math}_{g}\end{math} $\times$ M\begin{math}_{g}\end{math} $\times$ N\begin{math}_{g}\end{math} tensor. The Stage-II result s\begin{math}_{2}\end{math} is down-sampled to form M\begin{math}_{g}\end{math} $\times$ M\begin{math}_{g}\end{math} 
spatial size tensor. The image feature and the text feature tensors are concatenated along the channel dimension and the concatenated tensor is fed into several residual blocks which are designed to learn joint image and text features. This is followed by up-sampling to generate a W$\times$H high-resolution image. This generator aims to correct defects in the input image and simultaneously incorporates more details to increase the text relevancy.

\textbf{Inference :} The generated image inherits the high-resolution features of the second stage and has relatively more object features; however the added perceptual aspects are not significant compared with the cost of the increase in parameters.
Based on this preliminary research and the problems pointed out in DM-GAN paper~\cite{1}, we project that the presence of perceptual aspects should be addressed in the first stage of the GAN itself.

\section{\textbf{Results}}

\begin{figure*}
\centering
\includegraphics[width=18 cm]{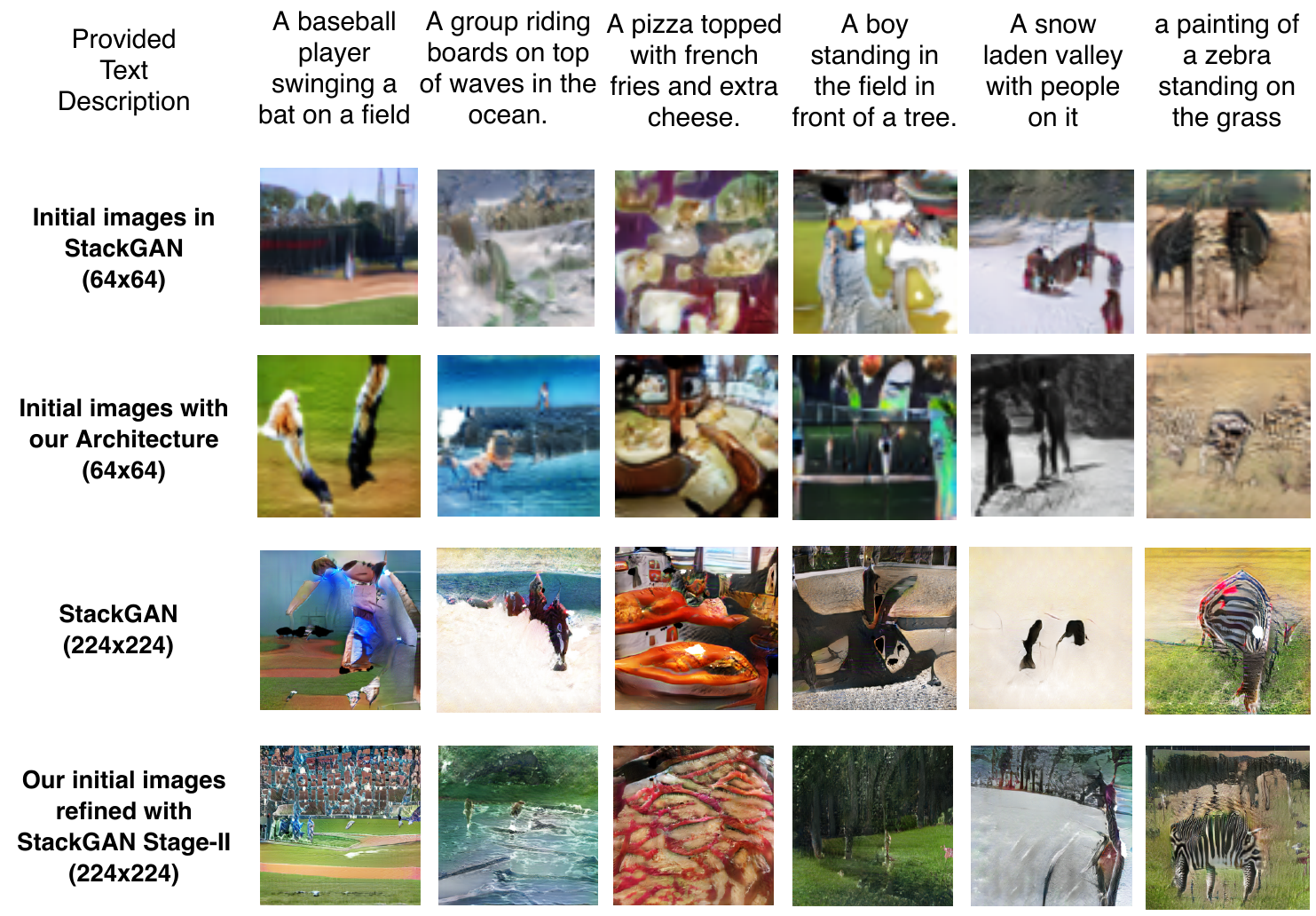}
\label{fig:image1}
\caption{Comparison of- Row 1: Generated initial images in StackGAN; Row 2: Initial images with our proposed PerceptionGAN; Row 3: Stage-II of StackGAN and; Row 4: Our initial image refined with StackGAN Stage-II; conditioned on text descriptions from MS-COCO test set.}
\end{figure*}

In \textbf{TABLE 1}, we compared our results with the state-of-the-art text to image generation methods on CUB, COCO datasets. Inception score~\cite{35} (a measure of how realistic a GAN's output is) is used as an evaluation metric. Although there is an increase in the inception score \textbf{(9.43 $\to$ 9.82)} from Stage-II to Stage-III, it is not very significant. Increasing the parameters doesn't significantly improve the generated image as the training data is limited. Improving the refinement stage~\cite{1,2} will definitely enhance the final generated image, but it is also equally important to improve the initial image generation. It is a must to account for the loss of perceptual features like shape, interactions, etc. in the initially generated image at Stage-I. We incorporated one such kind of loss into our PerceptionGAN architecture. In this paper, we integrated our novel PerceptionGAN architecture into the pipeline of StackGAN.  

It can be clearly seen that the inception score has increased significantly \textbf{(9.43 $\to$ 10.84)}, as shown in \textbf{TABLE 1} when our initial image is refined with StackGAN Stage-II.

\begin{table}[ht]
    \centering
    \begin{tabular}[H]{l| c c} 
\toprule
 \textbf{Methods}  & \textbf{CUB}  & \textbf{COCO}  \\[1.5 ex]
 \midrule
 GAN-INT-CLS~\cite{13}  & 2.88 $\pm$ .04 & 7.88 $\pm$ .07
 \\[2 ex]

 GAWWN~\cite{6}  & 3.62 $\pm$ .07 & /
 \\[2 ex]

 PPGN~\cite{7} &  / & 9.58 $\pm$ .21
 \\[2 ex]

 AttnGAN~\cite{2} & 4.36 $\pm$ .03 &  25.89 $\pm$ .47
 \\[2 ex]
 
 DM-GAN~\cite{1} & 4.75 $\pm$ .07 &  30.49 $\pm$ .57
 \\[2 ex]

 StackGAN~\cite{4}  & \textbf{3.70 $\pm$ .04} & \textbf{9.43 $\pm$ .03}
 \\[2 ex]

 StackGAN++~\cite{14} & 3.82 $\pm$ .06 & / 
 \\[2 ex]

  StackGAN with Added Stage-III &  3.86 $\pm$ 0.07 & 9.82 $\pm$ 0.13 
  \\[2 ex]

  StackGAN with Our Initial Image & \textbf{4.08 $\pm$ .09} & \textbf{10.84 $\pm$ .12}
 \\[2 ex]
  \bottomrule
    \end{tabular}
    \caption{Inception Scores (mean $\pm$ stdev) on CUB~\cite{23} and COCO~\cite{26} datasets}
    \label{tab:my_label}
\end{table}

Qualitative results are shown in \textbf{Fig. 6.}. It can be seen that the initial image generated with PerceptionGAN, when refined with StackGAN Stage-II, is relatively more interpretable than the output of the old StackGAN Stage-II image in terms of quality based on text relevancy. Some of the objects are still not properly generated in our output images because the efficient incorporation of the textual description in the refinement stage is equally important for generating realistic images. The StackGAN Stage-II (refinement used in this paper) models mainly the resolution quality of the image. Their Stage-II improvement is mostly related to color quality enhancement and not so much to perceptual information enhancement. This can be improved by enhancing the refinement stage, i.e., using the provided text description more efficiently into the refinement stage as done in DM-GAN~\cite{1} and AttnGAN~\cite{2} paper.

%-------------------------------------------------------------------

%-------------------------------------------------------------------
\section{\textbf{Conclusion}}
We proposed a novel architecture to incorporate perceptual understanding in the initial stage by adding Captioner loss in the discriminator module, which helps the generator to generate perceptually (shapes, colors, and object interactions) strong first stage images. These initially improved images will then be used in the refinement process by later stages to provide high-quality text conditioned images. Our proposed method can be integrated into the pipeline of state-of-the-art text-based-image-generation models such as DM-GAN and AttnGAN to generate initial low-resolution images. It is evident from the inception scores in \textbf{TABLE 1} and the qualitative results shown in \textbf{Fig. 6.} that image quality and relevancy have increased significantly when our initially generated images are refined with StackGAN Stage-II.
\\
The need to add the captioner loss is validated by a preliminary experiment on improving the refinement process in StackGAN by augmenting the third stage of the generator-discriminator pair in the StackGAN architecture. We derived from our experimental results shown in \textbf{TABLE 1} that adding more number of stages would not improve the generated image by a significant extent and neither will it lead to significant improvement in the inception score unless we account for the loss of perceptual features like shape, interactions, etc. in the initially generated image at Stage-I. We incorporated one such kind of loss in our architecture namely Captioner loss and shown significant improvement in generated image, qualitatively and quantitatively.

\subsection{\textbf{Future Work}}
We have applied our approach in StackGAN architecture in order to prove that strengthening the first stage for generating a base image in terms of perceptual information leads to significant improvement in the final generated image. A significant increment in the inception score is achieved when our initial generated image is refined with StackGAN Stage-II. We can very well suggest that if this approach is used with the current state-of-the-art, which is efficiently conditioning on the provided text and initially generated base image, results would be new state-of-the-art. 

\section*{Acknowledgment}
We want to express our special appreciation and thanks to our friends and colleagues for their contributions to this novel work. Computational resources are directly provided by Indian Institute of Technology Delhi. This work was partly supported by MOE Tier 2 grant no. MOE2018-T2-2-161 and SRG ISTD 2017 129.

\bibliographystyle{IEEEtran}
\bibliography{main}

\begin{thebibliography}{10}
\providecommand{\url}[1]{#1}
\csname url@samestyle\endcsname
\providecommand{\newblock}{\relax}
\providecommand{\bibinfo}[2]{#2}
\providecommand{\BIBentrySTDinterwordspacing}{\spaceskip=0pt\relax}
\providecommand{\BIBentryALTinterwordstretchfactor}{4}
\providecommand{\BIBentryALTinterwordspacing}{\spaceskip=\fontdimen2\font plus
\BIBentryALTinterwordstretchfactor\fontdimen3\font minus
  \fontdimen4\font\relax}
\providecommand{\BIBforeignlanguage}[2]{{%
\expandafter\ifx\csname l@#1\endcsname\relax
\typeout{** WARNING: IEEEtran.bst: No hyphenation pattern has been}%
\typeout{** loaded for the language `#1'. Using the pattern for}%
\typeout{** the default language instead.}%
\else
\language=\csname l@#1\endcsname
\fi
#2}}
\providecommand{\BIBdecl}{\relax}
\BIBdecl

\bibitem{28}
\BIBentryALTinterwordspacing
T.~Mikolov, K.~Chen, G.~Corrado, and J.~Dean, ``Efficient estimation of word
  representations in vector space,'' in \emph{1st International Conference on
  Learning Representations, {ICLR} 2013, Scottsdale, Arizona, USA, May 2-4,
  2013, Workshop Track Proceedings}, Y.~Bengio and Y.~LeCun, Eds., 2013.
  [Online]. Available: \url{http://arxiv.org/abs/1301.3781}
\BIBentrySTDinterwordspacing

\bibitem{29}
J.~Pennington, R.~Socher, and C.~D. Manning, ``Glove: Global vectors for word
  representation,'' in \emph{Proceedings of the 2014 conference on empirical
  methods in natural language processing (EMNLP)}, 2014, pp. 1532--1543.

\bibitem{30}
P.~Bojanowski, E.~Grave, A.~Joulin, and T.~Mikolov, ``Enriching word vectors
  with subword information,'' \emph{Transactions of the Association for
  Computational Linguistics}, vol.~5, pp. 135--146, 2017.

\bibitem{3}
I.~Goodfellow, J.~Pouget-Abadie, M.~Mirza, B.~Xu, D.~Warde-Farley, S.~Ozair,
  A.~Courville, and Y.~Bengio, ``Generative adversarial nets,'' in
  \emph{Advances in neural information processing systems}, 2014, pp.
  2672--2680.

\bibitem{4}
H.~Zhang, T.~Xu, H.~Li, S.~Zhang, X.~Wang, X.~Huang, and D.~N. Metaxas,
  ``Stackgan: Text to photo-realistic image synthesis with stacked generative
  adversarial networks,'' in \emph{Proceedings of the IEEE international
  conference on computer vision}, 2017, pp. 5907--5915.

\bibitem{1}
M.~Zhu, P.~Pan, W.~Chen, and Y.~Yang, ``Dm-gan: Dynamic memory generative
  adversarial networks for text-to-image synthesis,'' in \emph{Proceedings of
  the IEEE Conference on Computer Vision and Pattern Recognition}, 2019, pp.
  5802--5810.

\bibitem{2}
T.~Xu, P.~Zhang, Q.~Huang, H.~Zhang, Z.~Gan, X.~Huang, and X.~He, ``Attngan:
  Fine-grained text to image generation with attentional generative adversarial
  networks,'' in \emph{Proceedings of the IEEE conference on computer vision
  and pattern recognition}, 2018, pp. 1316--1324.

\bibitem{5}
E.~Mansimov, E.~Parisotto, J.~Ba, and R.~Salakhutdinov, ``Generating images
  from captions with attention,'' in \emph{ICLR}, 2016.

\bibitem{31}
S.~Reed, A.~van~den Oord, N.~Kalchbrenner, V.~Bapst, M.~Botvinick, and
  N.~De~Freitas, ``Generating interpretable images with controllable
  structure,'' 2016.

\bibitem{7}
A.~Nguyen, J.~Clune, Y.~Bengio, A.~Dosovitskiy, and J.~Yosinski, ``Plug \& play
  generative networks: Conditional iterative generation of images in latent
  space,'' in \emph{Proceedings of the IEEE Conference on Computer Vision and
  Pattern Recognition}, 2017, pp. 4467--4477.

\bibitem{13}
S.~Reed, Z.~Akata, X.~Yan, L.~Logeswaran, B.~Schiele, and H.~Lee, ``Generative
  adversarial text to image synthesis,'' in \emph{International Conference on
  Machine Learning}, 2016, pp. 1060--1069.

\bibitem{6}
S.~E. Reed, Z.~Akata, S.~Mohan, S.~Tenka, B.~Schiele, and H.~Lee, ``Learning
  what and where to draw,'' in \emph{Advances in neural information processing
  systems}, 2016, pp. 217--225.

\bibitem{8}
X.~Wang and A.~Gupta, ``Generative image modeling using style and structure
  adversarial networks,'' in \emph{European Conference on Computer
  Vision}.\hskip 1em plus 0.5em minus 0.4em\relax Springer, 2016, pp. 318--335.

\bibitem{9}
\BIBentryALTinterwordspacing
J.~Yang, A.~Kannan, D.~Batra, and D.~Parikh, ``{LR-GAN:} layered recursive
  generative adversarial networks for image generation,'' in \emph{5th
  International Conference on Learning Representations, {ICLR} 2017, Toulon,
  France, April 24-26, 2017, Conference Track Proceedings}.\hskip 1em plus
  0.5em minus 0.4em\relax OpenReview.net, 2017. [Online]. Available:
  \url{https://openreview.net/forum?id=HJ1kmv9xx}
\BIBentrySTDinterwordspacing

\bibitem{10}
X.~Huang, Y.~Li, O.~Poursaeed, J.~Hopcroft, and S.~Belongie, ``Stacked
  generative adversarial networks,'' in \emph{Proceedings of the IEEE
  conference on computer vision and pattern recognition}, 2017, pp. 5077--5086.

\bibitem{11}
I.~Durugkar, I.~Gemp, and S.~Mahadevan, ``Generative multi-adversarial
  networks,'' \emph{arXiv preprint arXiv:1611.01673}, 2016.

\bibitem{12}
E.~L. Denton, S.~Chintala, R.~Fergus \emph{et~al.}, ``Deep generative image
  models using a laplacian pyramid of adversarial networks,'' in \emph{Advances
  in neural information processing systems}, 2015, pp. 1486--1494.

\bibitem{14}
H.~Zhang, T.~Xu, H.~Li, S.~Zhang, X.~Wang, X.~Huang, and D.~N. Metaxas,
  ``Stackgan++: Realistic image synthesis with stacked generative adversarial
  networks,'' \emph{IEEE transactions on pattern analysis and machine
  intelligence}, vol.~41, no.~8, pp. 1947--1962, 2018.

\bibitem{39}
S.~Reed, Z.~Akata, H.~Lee, and B.~Schiele, ``Learning deep representations of
  fine-grained visual descriptions,'' in \emph{Proceedings of the IEEE
  Conference on Computer Vision and Pattern Recognition}, 2016, pp. 49--58.

\bibitem{25}
C.~Wang, H.~Yang, C.~Bartz, and C.~Meinel, ``Image captioning with deep
  bidirectional lstms,'' in \emph{Proceedings of the 24th ACM international
  conference on Multimedia}, 2016, pp. 988--997.

\bibitem{33}
\BIBentryALTinterwordspacing
M.~Arjovsky and L.~Bottou, ``Towards principled methods for training generative
  adversarial networks,'' in \emph{5th International Conference on Learning
  Representations, {ICLR} 2017, Toulon, France, April 24-26, 2017, Conference
  Track Proceedings}.\hskip 1em plus 0.5em minus 0.4em\relax OpenReview.net,
  2017. [Online]. Available: \url{https://openreview.net/forum?id=Hk4\_qw5xe}
\BIBentrySTDinterwordspacing

\bibitem{34}
\BIBentryALTinterwordspacing
A.~Radford, L.~Metz, and S.~Chintala, ``Unsupervised representation learning
  with deep convolutional generative adversarial networks,'' in \emph{4th
  International Conference on Learning Representations, {ICLR} 2016, San Juan,
  Puerto Rico, May 2-4, 2016, Conference Track Proceedings}, Y.~Bengio and
  Y.~LeCun, Eds., 2016. [Online]. Available:
  \url{http://arxiv.org/abs/1511.06434}
\BIBentrySTDinterwordspacing

\bibitem{35}
T.~Salimans, I.~Goodfellow, W.~Zaremba, V.~Cheung, A.~Radford, and X.~Chen,
  ``Improved techniques for training gans,'' in \emph{Advances in neural
  information processing systems}, 2016, pp. 2234--2242.

\bibitem{18}
\BIBentryALTinterwordspacing
T.~Karras, T.~Aila, S.~Laine, and J.~Lehtinen, ``Progressive growing of gans
  for improved quality, stability, and variation,'' in \emph{6th International
  Conference on Learning Representations, {ICLR} 2018, Vancouver, BC, Canada,
  April 30 - May 3, 2018, Conference Track Proceedings}.\hskip 1em plus 0.5em
  minus 0.4em\relax OpenReview.net, 2018. [Online]. Available:
  \url{https://openreview.net/forum?id=Hk99zCeAb}
\BIBentrySTDinterwordspacing

\bibitem{26}
T.-Y. Lin, M.~Maire, S.~Belongie, J.~Hays, P.~Perona, D.~Ramanan,
  P.~Doll{\'a}r, and C.~L. Zitnick, ``Microsoft coco: Common objects in
  context,'' in \emph{European conference on computer vision}.\hskip 1em plus
  0.5em minus 0.4em\relax Springer, 2014, pp. 740--755.

\bibitem{37}
K.~He, X.~Zhang, S.~Ren, and J.~Sun, ``Deep residual learning for image
  recognition,'' in \emph{Proceedings of the IEEE conference on computer vision
  and pattern recognition}, 2016, pp. 770--778.

\bibitem{23}
C.~Wah, S.~Branson, P.~Welinder, P.~Perona, and S.~Belongie, ``The caltech-ucsd
  birds-200-2011 dataset,'' 2011.

\end{thebibliography}
% \begin{thebibliography}{}
% \end{thebibliography}
\end{document}